\documentclass[twoside,11pt]{article}
\usepackage{jmlr2e}
\usepackage{wrapfig}
\usepackage{graphicx}
\usepackage{hyperref}
\usepackage{multirow}
\usepackage{color}

\title{Which Encoding is the Best for Text Classification in Chinese, English, Japanese and Korean?}

\author{\name Xiang Zhang \email xiang@cs.nyu.edu \\
  \addr Courant Institute of Mathematical Sciences, New York University \\
  \AND
  \name Yann LeCun \email yann@cs.nyu.edu \\
  \addr Courant Institute of Mathematical Sciences, New York University \\
  Center for Data Science, New York University \\
  Facebook AI Research, Facebook Inc.}

\editor{}

\begin{document}

\maketitle

\begin{abstract}
  This article offers an empirical study on the different ways of encoding Chinese, Japanese, Korean (CJK) and English languages for text classification. Different encoding levels are studied, including UTF-8 bytes, characters, words, romanized characters and romanized words. For all encoding levels, whenever applicable, we provide comparisons with linear models, fastText \citep{JGBM16} and convolutional networks. For convolutional networks, we compare between encoding mechanisms using character glyph images, one-hot (or one-of-n) encoding, and embedding. In total there are 473 models, using 14 large-scale text classification datasets in 4 languages including Chinese, English, Japanese and Korean. Some conclusions from these results include that byte-level one-hot encoding based on UTF-8 consistently produces competitive results for convolutional networks, that word-level n-grams linear models are competitive even without perfect word segmentation, and that fastText provides the best result using character-level n-gram encoding but can overfit when the features are overly rich.
\end{abstract}

\begin{keywords}
text classification, text encoding, text representation, multilingual language processing, convolutional network
\end{keywords}

\section{Introduction}

Being able to process different kinds of languages in a unified and consistent fashion is of great interest to the natural language processing (NLP) community, especially with the recent advancements in deep learning methods. Among these languages, Chinese, Japanese and Korean (CJK) pose unique challenges due to reasons in both linguistics and computation. Unlike some alphabetic languages such as English, there is no clear word boundary for some of the CJK texts. This makes it difficult to apply many laugnage processing methods that assume word as the basic construct.

Recently, many authors have proposed to use character-level encoding for language processing with convolutional networks (ConvNets) \citep{KJSR16} \citep{ZZL15}, casting away the word segmentation problem. Unfortunately, working with characters for CJK languages is not direct, because the amount of characters can be huge. For example, one-hot (or one-of-n) encoding used by \citet{ZZL15} is not practical because each one-hot vector would be prohibitively large.

\begin{wraptable}{r}{0.35\textwidth}
  \begin{center}
    \begin{tabular}{ll}
      \multicolumn{1}{c}{Layers}  &\multicolumn{1}{c}{Description}
      \\ \hline \\
      1-2           &Conv 256x3 \\
      3             &Pool 2 \\
      4-5           &Conv 256x3 \\
      6             &Pool 2 \\
      7-8           &Conv 256x3 \\
      9             &Pool 2 \\
      10-11         &Conv 256x3 \\
      12            &Pool 2 \\
      13-14         &Conv 256x3 \\
      15            &Pool 2 \\
      16-17         &Full 1024 \\
    \end{tabular}
  \end{center}
  \caption{The large classifier}
  \label{tab:lcls}
\end{wraptable}

This drives us to search for alternative ways of encoding CJK texts. The encoding mechanisms considered in this article include character glyph images, one-hot encoding and embedding. For one-hot encoding, we considered feasible encoding levels including UTF-8 bytes and characters after romanization. For embedding, we performed experiments on encoding levels including character, UTF-8 bytes, romanized characters, segmented word with a prebuilt word segmenter, and romanized word. A brief search in the literature seems to confirm that this article is the first to study all of these encoding mechanisms in a systematic fashion.

Historically, linear models such as (multinomial) logistic regression \citep{C58} and support vector machines \citep{CV95} have been the default choice for text classification, with bag-of-words features and variants such as n-grams and TF-IDF \citep{S72}. Therefore, in this article we provide extensive comparisons using multinomial logistic regression, with bag-of-characters, bag-of-words and their n-gram and TF-IDF \citep{S72} variants. Furthermore, experiments using the recently proposed fastText \citep{JGBM16} are also presented with all these different feature variants.

\begin{wraptable}{r}{0.35\textwidth}
  \begin{center}
    \begin{tabular}{ll}
      \multicolumn{1}{c}{Layers}  &\multicolumn{1}{c}{Description}
      \\ \hline \\
      1-2           &Conv 256x3 \\
      3             &Pool 3 \\
      4-5           &Conv 256x3 \\
      6             &Pool 3 \\
      7-8           &Conv 256x3 \\
      9             &Pool 3 \\
      10-11         &Full 1024 \\
    \end{tabular}
  \end{center}
  \caption{The small classifier}
  \label{tab:scls}
\end{wraptable}

Large-scale multi-lingual datasets are required to make sure that our comparisons are meaningful. Therefore, we set out to crawl the Internet for several large-scale text classification datasets. Eventually, we were able to obtain 14 large-scale datasets in 4 languages including Chinese, English, Japanese and Korean, for 2 different tasks including sentiment analysis and topic categorization. We plan to release all the code used in this article under an open source license, including crawling, preprocessing, and training on all datasets.

The conclusions of this article include that the one-hot encoding model at UTF-8 byte level consistently offers competitive results for convolutional networks, that linear models remain strong for the text classification task, and that fastText provides the best results with character n-grams but tends to overfit when the features are overly rich. We hope that these results can offer useful guidance for the community to select appropriate encoding mechanims that can handle different languages in a unified and consistent fashion.

\section{Encoding Mechanisms for Convolutional Networks}

For the purpose of fair comparisons, all of our convolutional networks share the same design except for the first few layers. We call this common part the classifier, and the different first several layers the encoder. In the benchmarks we have 2 classifier designs - one large and the other small. The large classifier consists of 12 layers, and the small one 8. Table \ref{tab:lcls} and \ref{tab:scls} details the designs. All parameterized layers use ReLU \citep{NH10} as the non-linearity.

\begin{wrapfigure}{r}{0.30\textwidth}
  \begin{center}
    \includegraphics[width=0.28\textwidth]{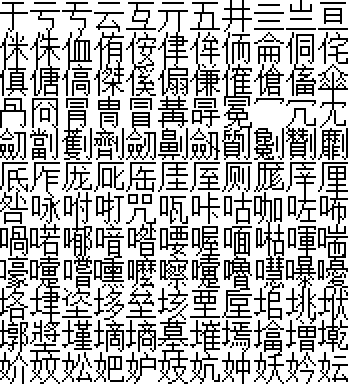}
  \end{center}
  \caption{GNU Unifont}
  \label{fig:unif}
\end{wrapfigure}

\subsection{Character Glyph}

Glyph is a typography term indicating a readable character for the purposes of writing. CJK languages consist of characters that are rich in their topological forms, where strokes and parts could represent semantic meaning. This makes glyph a potentially feasible encoding solution.

In the context of this article, we refer to glyphs as images of characters rendered by some font. In the experiments we use the freely available GNU Unifont \footnote{\url{http://unifoundry.com/unifont.html}} (version 8.0.01), where each character is converted to a 16-by-16 pixel image. We consider all characters that belong to the Unicode basic multi-lingual plane (BMP), which have code points less than or equal to the hex value FFFF. Figure \ref{fig:unif} shows some glyph examples in this font.

\begin{wraptable}{r}{0.4\textwidth}
  \begin{center}
    \begin{tabular}{ll}
      \multicolumn{1}{c}{Layers}  &\multicolumn{1}{c}{Description}
      \\ \hline \\
      1-2           &Conv 64x3x3 \\
      3             &Pool 2 \\
      4-5           &Conv 128x3x3 \\
      6             &Pool 2 \\
      7-8           &Conv 256x3x3 \\
      9             &Pool 2 \\
      10            &Full 1024 \\
      11            &Full 256
    \end{tabular}
  \end{center}
  \caption{Large GlyphNet encoder}
  \label{tab:lgen}
\end{wraptable}

For the large classifier, the glyph encoder contains 8 parameterized layers with 6 spatial convolutional layers and 2 linear layers. The small model consists of a 6-layer glyph encoder with 4 spatial convolutional layers and 2 linear layers. Table \ref{tab:lgen} and \ref{tab:sgen} present the design choices.

In the benchmarks we will refer to these 2 models as large GlyphNet and small GlyphNet respectively. During training, each sample consists of at most 512 characters for the large GlyphNet and 486 for the small one. Zero is padded if the length of the sample string is shorter, and characters beyond these limits are ignored. Note that each character must pass through the spatial glyph encoder and each sample could contain hundreds of characters. As a result, the training time of GlyphNet is significantly longer than any other model considered in this article.

It is worth noting that recent research has shown that CJK characters can help to improve the results of various tasks including text classification \citep{SKI16} \citep{LLLN17} and translation \citep{CAF17}, further justifying the potential of encoding CJK characters via glyphs.

\subsection{One-hot Encoding}

\begin{table}[t]
  \begin{center}
    \begin{tabular}{llcrrr}
      \multicolumn{1}{c}{Dataset}  & \multicolumn{1}{c}{Language}  & \multicolumn{1}{c}{Classes} & \multicolumn{1}{c}{Train} & \multicolumn{1}{c}{Test} & \multicolumn{1}{c}{Batch}
      \\ \hline \\
      Dianping & Chinese & 2 & 2,000,000 & 500,000 & 100,000 \\
      JD full & Chinese & 5 & 3,000,000 & 250,000 & 100,000 \\
      JD binary & Chinese & 2 & 4,000,000 & 360,000 & 100,000 \\
      Rakuten full & Japanese & 5 & 4,000,000 & 500,000 & 100,000 \\
      Rakuten binary & Japanese & 2 & 3,400,000 & 400,000 & 100,000 \\
      11st full & Korean & 5 & 750,000 & 100,000 & 100,000 \\
      11st binary & Korean & 2 & 4,000,000 & 400,000 & 100,000 \\
      Amazon full & English & 5 & 3,000,000 & 650,000 & 100,000 \\
      Amazon binary & English & 2 & 3,600,000 & 400,000 & 100,000 \\
      Ifeng & Chinese & 5 & 800,000 & 50,000 & 100,000 \\
      Chinanews & Chinese & 7 & 1,400,000 & 112,000 & 100,000 \\
      NYTimes & English & 7 & 1,400,000 & 105,000 & 100,000 \\
      Joint full & Multilingual & 5 & 10,750,000 & 1,500,000 & 400,000 \\
      Joint binary & Multilingual & 2 & 15,000,000 & 1,560,000 & 400,000 \\
    \end{tabular}
  \end{center}
  \caption{Datasets}
  \label{tab:data}
\end{table}

In the simplest version of one-hot (or one-of-n) encoding, each entity must be converted into a vector whose size equals to the cardinality of the set of all possible entities, and all values in this vector are zero except for the position that corresponds to the index of the entity in the set. For example, in the paper by \citet{ZZL15}, each entity is a character and the size of the vector equals to the size of the alphabet containing all characters. Unfortunately, this naive way of using one-hot encoding is only computationally feasible if the entity set is relatively small. Texts in CJK languages can easily span tens of thousands of characters.

\begin{wraptable}{r}{0.45\textwidth}
  \begin{center}
    \begin{tabular}{ll}
      \multicolumn{1}{c}{Layers}  &\multicolumn{1}{c}{Description}
      \\ \hline \\
      1-2           &Conv 64x3x3 \\
      3             &Pool 3 \\
      4-5           &Conv 128x3x3 \\
      6             &Pool 3 \\
      7-8           &Full 256
    \end{tabular}
  \end{center}
  \caption{Small GlyphNet encoder}
  \label{tab:sgen}
\end{wraptable}

In this article, we consider 2 simple solutions to this problem. The first one is to treat the text (in UTF-8) as a sequence of bytes and encode at byte-level. The second one, already presented in \citet{ZZL15}, is to romanize the text so that encoding using the English alphabet is feasible. Note that the second solution is equivalent of encoding at byte-level with romanized text, because the English alphabet is contained in UTF-8 and they will not go beyond the limit of one byte.

In the following we will call these 2 models byte-level OnehotNet and romanization OnehotNet. Similar to GlyphNet, each OnehotNet also has a large variant and a small variant depending on the classifier used. Both variants use the same encoder design that consists of 4 convolutional layers, in which the large variant admits input length 2048 and the small 1944. Table \ref{tab:ohen} provides the configuration. Compared to GlyphNet, OnehotNet is significantly faster because the encoder handles all symbols in the input at once.

The idea of language processing at byte level has been explored by \citet{GBVS16}, where they apply an LSTM-based \citep{HS97} sequence-to-sequence \citep{CMGBBSB14} \citep{SVL14} model at byte-level for a variety of tasks including part-of-speech tagging and named entity recognition, for 4 languages including English, German, Spanish and Dutch.  The advantage of byte-level processing is that they can be immediately applied to any language regardless of whether there are too many entities at character or word levels. The same advantage applies to CJK, and perhaps any language that can be digitized as well.

\subsection{Embedding}

\begin{wraptable}{r}{0.35\textwidth}
  \begin{center}
    \begin{tabular}{ll}
      \multicolumn{1}{c}{Layers}  &\multicolumn{1}{c}{Description}
      \\ \hline \\
      1-2           &Conv 256x3 \\
      3             &Pool 2 \\
      4-5           &Conv 256x3 \\
      6             &Pool 2 \\
    \end{tabular}
  \end{center}
  \caption{OnehotNet encoder}
  \label{tab:ohen}
\end{wraptable}

We use the terminology ``embedding'' to refer to the idea of associating each entity a fixed size vector, same as most papers in the machine learning literature. These vectors are randomly initialized, and then learnt either with an unsupervised criterion or jointly with the task at hand.  The advantage of embedding models is there there is no need to explicitly construct one-hot vectors, therefore the memory footprint of embedding models is significantly smaller than that of OnehotNet. As a result, embedding can be applied to almost any encoding level.

In this article, we use embedding at a variety of different levels, including byte, character, word, romanization character, and romanization word. All of of our emedding vectors are of size 256, and they are learnt jointly with the text classification task at hand. The size of vocabulary is 257 for byte-level and romanized-level encoding, 65537 for character-level encoding, and 200,002 for word level and romanized word-level encoding.

The character-level encoding considers all code points in the basic multilingual plane (BMP) of Unicode. The word and romanized-word vocabularies are built by selecting the 200,000 most frequent entities appeared in the training data for each dataset, plus one additional entry to represent an out-of-vocabulary symbol. One additional entry is also added to each vocabulary to include a padding symbol for shorter texts. There are 2 embedding models, since we have designed the classifier with 2 different sizes. We will refer to them as large EmbedNet and small EmbedNet respecitvely. The large Embednet admits input length of 512, and the small one 486.

When the input text is represented by explicit one-hot vectors, embedding is equivalent of using a linear first layer. Therefore, the difference between OnehotNet and EmbedNet in this article is whether the first layer is linear or convolutional. The idea of embedding has been applied to ConvNet-based text processing pretty early on, with representative work for tasks like named entity recognition, part-of-speech tagging\citep{CWBKKK11}, text classification at word level \citep{K14} and language modeling at character level \citep{KJSR16}.

\section{Linear Models and fastText}

Besides ConvNets, we also offer benchmarks in linear models using multinomial logistic regression, and the fastText program by \citet{JGBM16}.

\subsection{Linear Models}

The linear multinomial logistic regression models are all bag-of-entity models, where the entity is character, word, romanized word. The 1-gram bag-of-entity model admits a feature of size 200,000 by selecting the most frequent ones from the training dataset. The 5-gram model admis grams of length up to 5, using the 1,000,000 most frequent features in the training dataset.

\begin{wraptable}{r}{0.30\textwidth}
  \begin{center}
    \setlength\tabcolsep{2pt}
    \begin{tabular}{|l|r|r|}
      \hline
      \multicolumn{1}{|c|}{Dataset}  & \multicolumn{1}{c|}{Large} & \multicolumn{1}{c|}{Small} \\
      \hline
      Dianping & 24.31 & 24.52 \\
      JD f. & 48.97 & 49.29 \\
      JD b. & 9.85 & 10.08 \\
      Rakuten f. & 46.79 & 47.04 \\
      Rakuten b. & 6.65 & 6.83 \\
      11st f. & 32.72 & 33.01 \\
      11st b. & 13.89 & 14.30 \\
      Amazon f. & 46.62 & 47.92 \\
      Amazon b. & 8.54 & 9.10 \\
      Ifeng & 18.02 & 18.55 \\
      Chinanews & 12.26 & 12.89 \\
      NYTimes & 18.22 & 18.60 \\
      Joint f. & 45.19 & 45.80 \\
      Joint b. & 9.98 & 10.40 \\
      \hline
    \end{tabular}
  \end{center}
  \caption{GlyphNet results. The numbers are testing error in percentage.}
  \label{tab:glyr}
\end{wraptable}

Note that word segmentation is not a simple problem for some of CJK texts, because they sometimes do not contain clear word boundaries like the space character in most alphebatic languages. Section \ref{sect:wseg} introduces how word segmentation is done for each language.

The idea of bag-of-character and its n-gram version has been explored by \citet{PHSW03} for text classification in Asian languages, where they observed comparable results with word-level models. This is probably because of the large character vocabularies in these languages, in which each character has a similar sparsity in representing meaning compared to each word in an alphebetic language.

\subsection{fastText}

fastText \citep{JGBM16} is a recent tool for fast text classification by incorporating several tricks such as hierarchical softmax \citep{G01} \citep{MCCD13} and feature hashing \citep{WDLSA09}. Combined with an efficient implementation and a highly optimized learning rate schedule, fastText is able to process input text at a speed of several orders of magnitude of that of ConvNets. This gives it a particular advantage and we hope to include the its results as a reference for our community.

The fastText model is essentially a 2-layer fully connected neural network without non-linearity. The number of hidden units is 10 across all of our experiments. During training, we use an initial learning rate of 0.1 and a hashing bucket size of 10,000,000. We used 10\% of the training dataset as validation and remaining as training to choose the best number of epoches, from the choices 2, 5 and 10. This validation process necessary because fastText does not have weight decay \citep{JGBM16} and it relies on early stopping to prevent overfitting. It is also the only model fast enough for such hyper-parameter tuning in this article. For each dataset, we explored features at character, word and romanized word levels, with variants of 1-gram, 2-gram and 5-gram features.

\section{Datasets and Preprocessing}

To ensure that our results are significant enough to demonstrate the differences between encoding methods, we need to acquire large-scale datasets. To do that, we set out to crawl the Internet for text classification datasets in 4 language including Chinese, English, Japanese and Korean. Eventually, we were able to obtain 14 datasets, most of which are at the scale of millions of samples. We performed experiments using all aforementioned models on all of these datasets.

\subsection{Datasets}

In total, we have obtained 8 sentiment classification datasets from online shopping reviews in Chinese, English, Japanese and Korean, 1 sentiment classification dataset from online restaurant reviews in Chinese, and 3 news topic classification dataset in Enlish and Chinese. Additionally, we were able to combine the online shopping review datasets in different languages to construct 2 joint datasets, which can be used to test each model's ability to handle different languages in a unified fashion. Table \ref{tab:data} summarizes the statistics of all these datasets.

\begin{wraptable}{r}{0.45\textwidth}
  \begin{center}
    \setlength\tabcolsep{2pt}
    \begin{tabular}{|l|r|r|r|r|}
      \hline
      \multicolumn{1}{|c|}{\multirow{2}{*}{Dataset}}  & \multicolumn{2}{c|}{Byte} & \multicolumn{2}{c|}{Romanized} \\ \cline{2-5}
      & \multicolumn{1}{c|}{large} & \multicolumn{1}{c|}{small} & \multicolumn{1}{c|}{large} & \multicolumn{1}{c|}{small} \\
      \hline
      Dianping & \textcolor{blue}{\textbf{23.17}} & 23.31 & 23.53 & \textcolor{red}{\textbf{23.72}} \\
      JD f. & \textcolor{blue}{\textbf{48.10}} & 48.27 & 48.42 & \textcolor{red}{\textbf{48.53}} \\
      JD b. & 9.33 & \textcolor{blue}{\textbf{9.31}} & 9.49 & \textcolor{red}{\textbf{9.51}} \\
      Rakuten f. & \textcolor{blue}{\textbf{45.10}} & 45.38 & 45.13 & \textcolor{red}{\textbf{45.38}} \\
      Rakuten b. & \textcolor{blue}{\textbf{5.93}} & \textcolor{red}{\textbf{6.10}} & 6.03 & 6.08 \\
      11st f. & 32.56 & \textcolor{blue}{\textbf{32.43}} & \textcolor{red}{\textbf{32.73}} & 32.69 \\
      11st b. & \textcolor{blue}{\textbf{13.30}} & 13.33 & 13.41 & \textcolor{red}{\textbf{13.45}} \\
      Amazon f. & \textcolor{blue}{\textbf{42.21}} & \textcolor{red}{\textbf{42.31}} & \multicolumn{1}{|c|}{--} & \multicolumn{1}{|c|}{--} \\
      Amazon b. & \textcolor{blue}{\textbf{6.52}} & \textcolor{red}{\textbf{6.59}} &  \multicolumn{1}{|c|}{--} & \multicolumn{1}{|c|}{--} \\
      Ifeng & 16.70 & \textcolor{blue}{\textbf{16.49}} & \textcolor{red}{\textbf{18.92}} & 18.90 \\
      Chinanews & \textcolor{blue}{\textbf{10.62}} & 10.73 & 11.71 & \textcolor{red}{\textbf{11.76}} \\
      NYTimes & \textcolor{red}{\textbf{14.30}} & \textcolor{blue}{\textbf{14.26}} & \multicolumn{1}{|c|}{--} & \multicolumn{1}{|c|}{--} \\
      Joint f. & \textcolor{blue}{\textbf{42.93}} & 43.09 & \textcolor{red}{\textbf{43.29}} & 43.26 \\
      Joint b. & 8.79 & \textcolor{blue}{\textbf{8.78}} & 9.00 & \textcolor{red}{\textbf{9.02}} \\
      \hline
    \end{tabular}
  \end{center}
  \caption{OnehotNet results. The numbers are testing error in percentage. The best result for each dataset is marked blue and the worst red.}
  \label{tab:oner}
\end{wraptable}

\textbf{Dianping}. The Dianping dataset consists of user reviews crawled from Chinese online restaurant review website \url{dianping.com}. This dataset was developed and used by Zhang et al. for research in collaborative filtering \citep{ZZLM13} \citep{ZZLMF13} and sentiment analysis \citep{ZLZZLM14} \citep{ZZZLM14}. After removing duplicated texts, we preprocessed the dataset such that stars 1, 2 and 3 belong to the negative class, and stars 4 and 5 belong to the positive class. Then we randomly selected 2,000,000 samples for training and 500,000 samples for testing with equal number of samples in each sentiment.

\textbf{JD}. The JD dataset consists of user reviews crawled from the Chinese online shopping website \url{jd.com}. After duplication removal, we were able to obtain 2 sentiment classification datasets in which one is to predict the full 5 stars and the other is binary. The binary dataset was built such that stars 1 and 2 belong to the negative sentiment, and stars 4 and 5 belong to the positive sentiment. Star 3 is ignored in the JD binary dataset. There are 3,000,000 training samples and 250,000 testing samples in the JD full dataset, and 4,000,000 training samples and 360,000 testing samples in the JD binary dataset. In each case, the samples are evenly distributed across classes.

\textbf{Rakuten}. The Rakuten dataset consists of user reviews cralwed from the Japanese online shopping webiste \url{rakuten.co.jp}. After duplication removal, we were able to obtain 2 sentiment classification datasets in which one is to predict the full 5 stars and the other is binary. The binary dataset was built such that stars 1 and 2 belong to the negative sentiment, and stars 4 and 5 belong to the positive sentiment. Star 3 is ignored in the Rakuten binary dataset. There are 4,000,000 training samples and 500,000 testing samples in the Rakuten full dataset, and 3,400,000 training samples and 400,000 testing samples in the Rakuten binary dataset. In each case, the samples are evenly distributed across classes.

\begin{table}[t]
  \begin{center}
    \setlength\tabcolsep{2pt}
    \begin{tabular}{|l|r|r|r|r|r|r|r|r|r|r|}
      \hline
      \multicolumn{1}{|c|}{\multirow{2}{*}{Dataset}} & \multicolumn{2}{c|}{Character} & \multicolumn{2}{c|}{Byte} & \multicolumn{2}{c|}{Romanized} & \multicolumn{2}{c|}{Word} & \multicolumn{2}{c|}{Rom. word} \\ \cline{2-11}
      & \multicolumn{1}{c|}{large} & \multicolumn{1}{c|}{small} & \multicolumn{1}{c|}{large} & \multicolumn{1}{c|}{small} & \multicolumn{1}{c|}{large} & \multicolumn{1}{c|}{small} & \multicolumn{1}{c|}{large} & \multicolumn{1}{c|}{small} & \multicolumn{1}{c|}{large} & \multicolumn{1}{c|}{small} \\
      \hline
      Dianping & \textcolor{blue}{\textbf{23.60}} & 23.67 & 24.09 & 24.22 & 25.42 & \textcolor{red}{\textbf{26.04}} & 24.55 & 24.69 & 23.70 & 23.81 \\
      JD f. & \textcolor{blue}{\textbf{48.29}} & 48.43 & 48.56 & 48.66 & 48.75 & 49.28 & 50.05 & \textcolor{red}{\textbf{50.14}} & 49.15 & 49.25 \\
      JD b. & 9.43 & 9.41 & \textcolor{blue}{\textbf{9.19}} & 9.21 & 9.46 & 9.71 & 10.37 & \textcolor{red}{\textbf{10.44}} & 9.58 & 9.69 \\
      Rakuten f. & \textcolor{blue}{\textbf{45.20}} & 45.68 & 45.96 & \textcolor{red}{\textbf{46.89}} & 46.15 & 46.78 & 46.34 & 46.55 & 45.96 & 46.34 \\
      Rakuten b. & \textcolor{blue}{\textbf{6.07}} & 6.13 & 6.49 & 6.86 & 6.56 & \textcolor{red}{\textbf{6.95}} & 6.81 & 6.86 & 6.55 & 6.67 \\
      11st f. & \textcolor{blue}{\textbf{32.29}} & 32.34 & 34.84 & 35.02 & 35.43 & 35.70 & 37.71 & 37.52 & \textcolor{red}{\textbf{42.60}} & 42.53 \\
      11st b. & 13.43 & 13.50 & \textcolor{blue}{\textbf{13.25}} & 13.47 & 13.48 & 13.70 & 15.16 & 14.26 & 17.65 & \textcolor{red}{\textbf{17.68}} \\
      Amazon f. & \textcolor{blue}{\textbf{43.70}} & 44.22 & \multicolumn{1}{c|}{--} & \multicolumn{1}{c|}{--} & \multicolumn{1}{c|}{--} & \multicolumn{1}{c|}{--} & 44.26 & \textcolor{red}{\textbf{44.82}} & \multicolumn{1}{c|}{--} & \multicolumn{1}{c|}{--} \\
      Amazon b. & \textcolor{blue}{\textbf{7.17}} & 7.48 & \multicolumn{1}{c|}{--} & \multicolumn{1}{c|}{--} & \multicolumn{1}{c|}{--} & \multicolumn{1}{c|}{--} & 7.92 & \textcolor{red}{\textbf{8.02}} & \multicolumn{1}{c|}{--} & \multicolumn{1}{c|}{--} \\
      Ifeng & \textcolor{blue}{\textbf{17.01}} & 17.08 & 17.12 & 17.55 & 19.21 & 20.00 & \textcolor{red}{\textbf{20.82}} & 20.73 & 19.46 & 19.48 \\
      Chinanews & 11.04 & 11.12 & \textcolor{blue}{\textbf{10.55}} & 10.84 & 11.84 & 12.77 & 14.75 & \textcolor{red}{\textbf{14.95}} & 11.92 & 12.06 \\
      NYTimes & \textcolor{blue}{\textbf{14.12}} & 14.60 & \multicolumn{1}{c|}{--} & \multicolumn{1}{c|}{--} & \multicolumn{1}{c|}{--} & \multicolumn{1}{c|}{--} & 17.63 & \textcolor{red}{\textbf{17.79}} & \multicolumn{1}{c|}{--} & \multicolumn{1}{c|}{--} \\
      Joint f. & \textcolor{blue}{\textbf{43.64}} & 44.11 & 44.19 & 44.79 & 44.74 & 45.46 & 45.01 & 45.31 & 45.02 & \textcolor{red}{\textbf{45.34}} \\
      Joint b. & \textcolor{blue}{\textbf{9.02}} & 9.18 & 9.09 & 9.27 & 9.34 & 9.65 & \textcolor{red}{\textbf{10.75}} & 10.01 & 9.94 & 10.02 \\
      \hline
    \end{tabular}
  \end{center}
  \caption{EmbedNet results. The numbers are testing error in percentage. The best result for each dataset is marked blue and the worst red.}
  \label{tab:embr}
\end{table}

\textbf{11st}. The 11st dataset consists of user reviews crawled from the Korean online shopping website \url{11st.co.kr}. After duplication removal, we were able to obtain 2 sentiment classification datasets in which one is to predict the full 5 stars and the other is binary. The binary dataset was built such that stars 1, 2 and 3 belong to the negative sentiment, and stars 4 and 5 belong to the positive sentiment. There are 750,000 training samples and 100,000 testing samples in the 11st full dataset, and 4,000,000 training samples and 400,000 testing samples in the 11st binary dataset. In each case, the samples are evenly distributed across classes.

\textbf{Amazon}. The Amazon dataset consists of users reviews crawled from the English online shopping website \url{amazon.com}. We use the same datasets constructed by \citet{ZZL15}, which came from the Stanford Network Analysis Project (SNAP) \footnote{\url{http://snap.stanford.edu/}} and developed by \citet{ML13} for sentiment analysis. There are 2 sentiment classification datasets in which one is to predict the full 5 stars and the other is binary. The binary dataset was built such that stars 1 and 2 belong to the negative sentiment, and stars 4 and 5 belong to the positive sentiment. Star 3 is ignored in the Amazon binary dataset. There are 3,000,000 training samples and 650,000 testing samples in the Amazon full dataset, and 3,600,000 training samples and 400,000 testing samples in the Amazon binary dataset. In each case, the samples are evenly distributed across classes.

\textbf{Ifeng}. The Ifeng dataset consists of first paragraphs of news articles from the Chinese news website \url{ifeng.com}. We crawled all news from the year 2006 to the year 2016 and selected 5 different news channels as 5 topic classes. These classes are mainland China politics, International news, Taiwan - Hong Kong- Macau politics, military news, and society news. After duplication removal, the dataset consists of 800,000 training samples and 50,000 testing samples. These samples are evenly distributed across classes.

\begin{table}[t]
  \begin{center}
    \setlength\tabcolsep{2pt}
    \begin{tabular}{|l|r|r|r|r|r|r|r|r|r|r|r|r|}
      \hline
      \multicolumn{1}{|c|}{\multirow{3}{*}{Dataset}}  & \multicolumn{4}{c|}{Character} & \multicolumn{4}{c|}{Word} & \multicolumn{4}{c|}{Romanized Word} \\ \cline{2-13}
      & \multicolumn{2}{c|}{1-gram} & \multicolumn{2}{c|}{5-gram} & \multicolumn{2}{c|}{1-gram} & \multicolumn{2}{c|}{5-gram} & \multicolumn{2}{c|}{1-gram} & \multicolumn{2}{c|}{5-gram} \\ \cline{2-13}
      & \multicolumn{1}{c|}{plain} & \multicolumn{1}{c|}{tfidf} & \multicolumn{1}{c|}{plain} & \multicolumn{1}{c|}{tfidf} & \multicolumn{1}{c|}{plain} & \multicolumn{1}{c|}{tfidf} & \multicolumn{1}{c|}{plain} & \multicolumn{1}{c|}{tfidf} & \multicolumn{1}{c|}{plain} & \multicolumn{1}{c|}{tfidf} & \multicolumn{1}{c|}{plain} & \multicolumn{1}{c|}{tfidf} \\
      \hline
      Dianping & 26.07 & 26.82 & 24.29 & 23.59 & 24.05 & 24.26 & 23.56 & \textcolor{blue}{\textbf{23.03}} & 27.32 & \textcolor{red}{\textbf{28.02}} & 24.54 & 23.35 \\
      JD f. & 51.47 & 51.63 & 48.43 & \textcolor{blue}{\textbf{48.18}} & 49.39 & 50.06 & 48.30 & 48.30 & 52.14 & \textcolor{red}{\textbf{52.77}} & 48.97 & 48.47 \\
      JD b. & 11.81 & 12.13 & 9.10 & 8.92 & 9.86 & 9.98 & 9.07 & \textcolor{blue}{\textbf{8.82}} & 13.11 & \textcolor{red}{\textbf{13.53}} & 9.12 & 8.98 \\
      Rakuten f. & 52.15 & \textcolor{red}{\textbf{52.82}} & 47.62 & 46.47 & 47.57 & 47.73 & 45.79 & \textcolor{blue}{\textbf{45.26}} & 47.91 & 48.31 & 46.97 & 45.66 \\
      Rakuten b. & 12.52 & \textcolor{red}{\textbf{13.00}} & 8.15 & 7.30 & 8.38 & 8.40 & 7.34 & \textcolor{blue}{\textbf{6.63}} & 8.82 & 8.95 & 7.46 & 6.72 \\
      11st f. & 43.87 & \textcolor{red}{\textbf{48.35}} & 43.59 & \textcolor{blue}{\textbf{43.42}} & 45.30 & 45.05 & 43.86 & 43.44 & 45.37 & 44.77 & 45.56 & 44.23 \\
      11st b. & 17.66 & \textcolor{red}{\textbf{18.01}} & 14.45 & 14.35 & 15.34 & 15.59 & 13.73 & \textcolor{blue}{\textbf{13.40}} & 14.96 & 15.16 & 14.62 & 14.43 \\
      Amazon f. & \textcolor{red}{\textbf{69.47}} & 68.60 & 56.86 & 51.33 & 45.39 & 44.90 & 44.67 & \textcolor{blue}{\textbf{42.70}} & \multicolumn{1}{c|}{--} & \multicolumn{1}{c|}{--} & \multicolumn{1}{c|}{--} & \multicolumn{1}{c|}{--} \\
      Amazon b. & \textcolor{red}{\textbf{34.47}} & 33.69 & 15.00 & 12.15 & 9.33 & 8.80 & 8.55 & \textcolor{blue}{\textbf{8.23}} & \multicolumn{1}{c|}{--} & \multicolumn{1}{c|}{--} & \multicolumn{1}{c|}{--} & \multicolumn{1}{c|}{--} \\
      Ifeng & 22.12 & 22.43 & 21.52 & 21.98 & 19.12 & \textcolor{blue}{\textbf{18.30}} & 20.17 & 19.73 & 26.59 & \textcolor{red}{\textbf{27.33}} & 23.12 & 22.38 \\
      Chinanews & 15.35 & 15.07 & 14.92 & 13.37 & 11.63 & \textcolor{blue}{\textbf{10.76}} & 13.42 & 12.92 & 20.04 & \textcolor{red}{\textbf{20.48}} & 15.56 & 13.98 \\
      NYTimes & \textcolor{red}{\textbf{57.55}} & 53.86 & 40.91 & 26.41 & 18.24 & \textcolor{blue}{\textbf{15.31}} & 20.05 & 18.30 & \multicolumn{1}{c|}{--} & \multicolumn{1}{c|}{--} & \multicolumn{1}{c|}{--} & \multicolumn{1}{c|}{--} \\
      Joint f. & \textcolor{red}{\textbf{60.28}} & 59.69 & 49.19 & 48.21 & 46.82 & 46.54 & 45.26 & \textcolor{blue}{\textbf{45.07}} & 47.53 & 47.16 & 46.89 & 46.39 \\
      Joint b. & \textcolor{red}{\textbf{20.20}} & 19.73 & 12.12 & 10.91 & 10.83 & 10.66 & 9.46 & \textcolor{blue}{\textbf{9.00}} & 11.73 & 11.44 & 11.31 & 11.01 \\ \hline
    \end{tabular}
  \end{center}
  \caption{Linear model results. The numbers are testing error in percentage. The best result for each dataset is marked blue and the worst red.}
  \label{tab:linr}
\end{table}

\textbf{Chinanews}. The Chinanews daaset consists of first paragraphs of news articles from the Chinese news website \url{chinanews.com}. We crawled all news from the year 2008 to the year 2016 and selected 7 different news channels as 7 topic classes. These classes are mainland China politics, Hong Kong - Macau politics, Taiwan politics, International news, financial news, culture, entertainment, sports, and health. After duplication removal, the dataset consists of 1,400,000 training samples and 112,000 testing samples. These samples are evenly distributed across classes.

\textbf{NYTimes}. The NYTimes dataset consists of first paragraphs of news articles from the English news website \url{nytimes.com}. We crawled all news from the year 1981 to the year 2015 and combined several channels to construct 7 topic classes. These classes are business news, New York regional news, sports, U.S. politics, world news and opinions, arts and fashion, and entertainment and science. After duplication removal, the dataset consists of 1,400,000 training samples and 105,000 testing samples. These samples are evenly distributed acorss classes.

\begin{table}[t]
  \begin{center}
    \setlength\tabcolsep{2pt}
    \begin{tabular}{|l|r|r|r|r|r|r|r|r|r|}
      \hline
      \multicolumn{1}{|c|}{\multirow{2}{*}{Dataset}} & \multicolumn{3}{c|}{Character} & \multicolumn{3}{c|}{Word} & \multicolumn{3}{c|}{Romanized Word} \\ \cline{2-10}
      & \multicolumn{1}{c|}{1-gram} & \multicolumn{1}{c|}{2-gram} & \multicolumn{1}{c|}{5-gram} & \multicolumn{1}{c|}{1-gram} & \multicolumn{1}{c|}{2-gram} & \multicolumn{1}{c|}{5-gram} & \multicolumn{1}{c|}{1-gram} & \multicolumn{1}{c|}{2-gram} & \multicolumn{1}{c|}{5-gram} \\
      \hline
      Dianping & 25.83 & 22.85 & \textcolor{blue}{\textbf{22.34}} & 23.72 & 22.62 & 22.62 & \textcolor{red}{\textbf{26.99}} & 22.90 & 22.42 \\
      JD f. & 51.25 & 48.29 & \textcolor{blue}{\textbf{47.99}} & 49.23 & 48.11 & 48.59 & \textcolor{red}{\textbf{52.21}} & 48.36 & 48.13 \\
      JD b. & 11.81 & 8.91 & \textcolor{blue}{\textbf{8.72}} & 9.84 & 9.11 & 9.93 & \textcolor{red}{\textbf{13.04}} & 9.06 & 8.73 \\
      Rakuten f. & \textcolor{red}{\textbf{51.98}} & 44.89 & \textcolor{blue}{\textbf{43.27}} & 46.57 & 43.81 & 46.31 & 47.06 & 43.79 & 43.28 \\
      Rakuten b. & \textcolor{red}{\textbf{12.24}} & 6.93 & \textcolor{blue}{\textbf{5.45}} & 8.03 & 5.90 & 5.45 & 8.47 & 5.94 & 5.44 \\
      11st f. & \textcolor{red}{\textbf{43.20}} & 38.70 & \textcolor{blue}{\textbf{38.58}} & 40.73 & 38.65 & 38.67 & 40.71 & 41.73 & 43.16 \\
      11st b. & \textcolor{red}{\textbf{17.66}} & 13.65 & \textcolor{blue}{\textbf{13.11}} & 15.36 & 13.48 & 13.23 & 14.87 & 14.51 & 14.97 \\
      Amazon f. & \textcolor{red}{\textbf{67.07}} & 53.99 & 41.05 & 43.81 & 40.20 & \textcolor{blue}{\textbf{40.02}} & \multicolumn{1}{c|}{--} & \multicolumn{1}{c|}{--} & \multicolumn{1}{c|}{--} \\
      Amazon b. & \textcolor{red}{\textbf{32.79}} & 18.57 & 6.28 & 8.37 & 5.59 & \textcolor{blue}{\textbf{5.41}} & \multicolumn{1}{c|}{--} & \multicolumn{1}{c|}{--} & \multicolumn{1}{c|}{--} \\
      Ifeng & \textcolor{red}{\textbf{21.60}} & 16.58 & \textcolor{blue}{\textbf{16.31}} & 17.82 & 16.65 & 16.95 & 26.04 & 18.21 & 17.86 \\
      Chinanews & 13.92 & 9.31 & \textcolor{blue}{\textbf{9.10}} & 9.86 & 9.25 & 9.24 & \textcolor{red}{\textbf{18.68}} & 9.66 & 9.39 \\
      NYTimes & \textcolor{red}{\textbf{51.42}} & 24.67 & 12.72 & 13.60 & \textcolor{blue}{\textbf{11.84}} & 13.23 & \multicolumn{1}{c|}{--} & \multicolumn{1}{c|}{--} & \multicolumn{1}{c|}{--} \\
      Joint f. & \textcolor{red}{\textbf{59.49}} & 49.37 & 44.03 & 46.03 & 43.36 & 43.29 & 46.72 & \textcolor{blue}{\textbf{43.14}} & 43.26 \\
      Joint b. & \textcolor{red}{\textbf{19.85}} & 12.35 & 9.00 & 10.57 & \textcolor{blue}{\textbf{8.65}} & 8.74 & 11.39 & 8.81 & 8.84 \\
      \hline
    \end{tabular}
  \end{center}
  \caption{fastText results. The numbers are testing error in percentage. The best result for each dataset is marked blue and the worst red.}
  \label{tab:fstr}
\end{table}

\textbf{Joint}. The four dataset sources JD, Rakuten, 11st and Amazon are all sentiment classification tasks from online shopping websites, with both full 5 stars prediction or binary prediction. Therefore, we could combine them in each case to form two new joint datasets of 5 classes or 2 classes. This dataset is particularly useful since it spans 4 languages and can be used to test a model's ability to handle different languages in a unified fashion. In total, there are 10,750,000 trainig samples and 1,500,000 testing samples in the joint full dataset, and 15,000,000 training samples and 1,560,000 testing samples in the joint binary dataset. All samples are evenly distributed across classes.

\subsection{Word Segmentation and Romanization}
\label{sect:wseg}

Since there is no clear word boundary in some of the CJK texts, word segmentation is necessary before applying any of the word-level models. Romanization for some of the CJK texts also depends on word segmentation to produce the correct transliteration in the English alphabet. In this section, we present both word segmentation and romanization processes used for producing the results, for each languages Chinese, Japanese and Korean. All the tools we used are relatively popular and standard for CJK language processing.

\textbf{Chinese}. For Chinese, we use the freely available word segmentation package called jieba \footnote{\url{https://github.com/fxsjy/jieba}} (version 0.38). The romanization standard we used is Pinyin, using the pypinyin \footnote{\url{https://github.com/mozillazg/python-pinyin}} (version 0.12) package which in turn calls jieba for disambiguate between characters with multiple pronunciations.

\textbf{Japanese}. For Japanese, we use the freely available word segmentation and tagging package MeCab \footnote{\url{http://taku910.github.io/mecab}} (version 0.996) with the default model for Japanese. The romanization form used is Hepburn, which is done by converting the segmented words using python-romkan \footnote{\url{https://www.soimort.org/python-romkan}} (version 0.2.1).

\begin{figure}[t]
  \begin{center}
    \includegraphics[width=\textwidth]{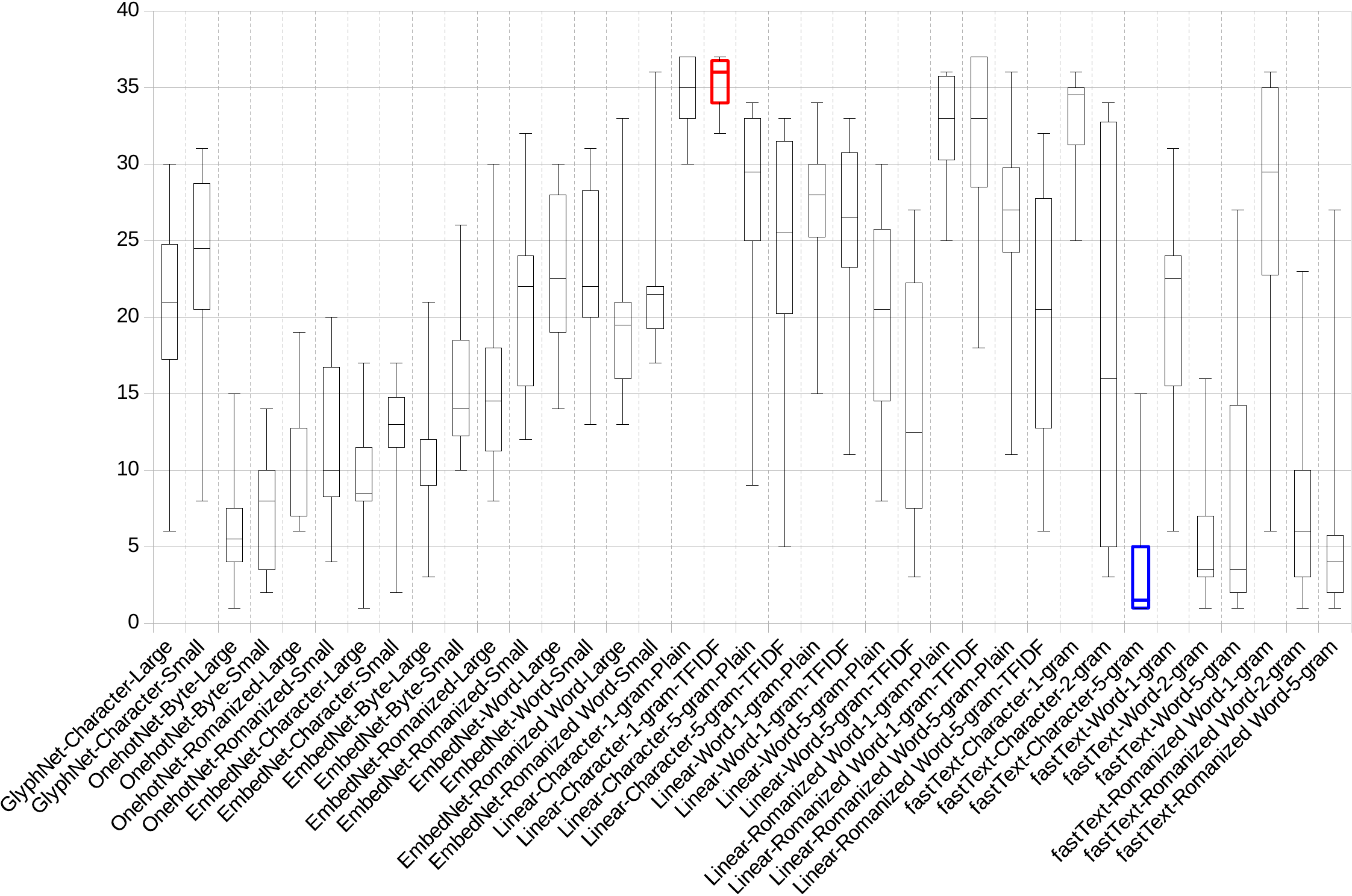}
  \end{center}
  \caption{Rank box plot of testing error for different models}
  \label{fig:rank}
\end{figure}

\textbf{Korean}. Word segmentation is done for Korean using MeCab as well, but with a model in the Korean language \footnote{\url{https://bitbucket.org/eunjeon/mecab-ko-dic}}. Instead of calling MeCab and parsing the results like that in Japanese, we used the MeCab wrapper in KoNLPy \footnote{\url{http://konlpy.org}} which offers rich information for Korean text. The romanization standard used is the Revised Romanization of Korean (RR), which is done in 2 steps. The first step is to convert any Hanja in the text to Hangul via the python package hanja (version 0.11) \footnote{\url{https://github.com/suminb/hanja}}, and the second step is to transliterate the generate Hangul using the python package hangul-romanize \footnote{\url{https://github.com/youknowone/hangul-romanize}}.

\section{Experiments}

After introducing the optimization parameters used for all of our models, this section then presents the results for these models. Most of our experiments are implemented using Torch 7 \citep{CKF11}, with NVIDIA CUDNN \footnote{\url{https://developer.nvidia.com/cudnn}} as the GPU backend.

\subsection{Optimization}

The optimization process used for all convolutional network models is stochastic gradient descent (SGD) with momentum \citep{P64} \citep{SMDH13}. The training process operates on random minimabatches of size 16, with different numbers of minibatches per epoch. The sixth column in Table \ref{tab:data} shows the number of minibatches for one epoch for each dataset. The model parameters are initialized in the same way as in \citet{HZRS15} -- for each layer the bias is set to 0, and weights are randomly sampled from a Gaussian distribution of mean 0 and standard deviation \(\sqrt{2/n}\), where \(n\) is the number of output units each input unit connects to. All the models have an initial learning rate of 0.00001, which is halved every 8 epoches. The training stops at the 100th epoch. A small weight decay of 0.00001 is applied to the model to stabilize training. Each model is trained using one NVIDIA Tesla K40 GPU.

\begin{figure}[t]
  \begin{center}
    \includegraphics[width=\textwidth]{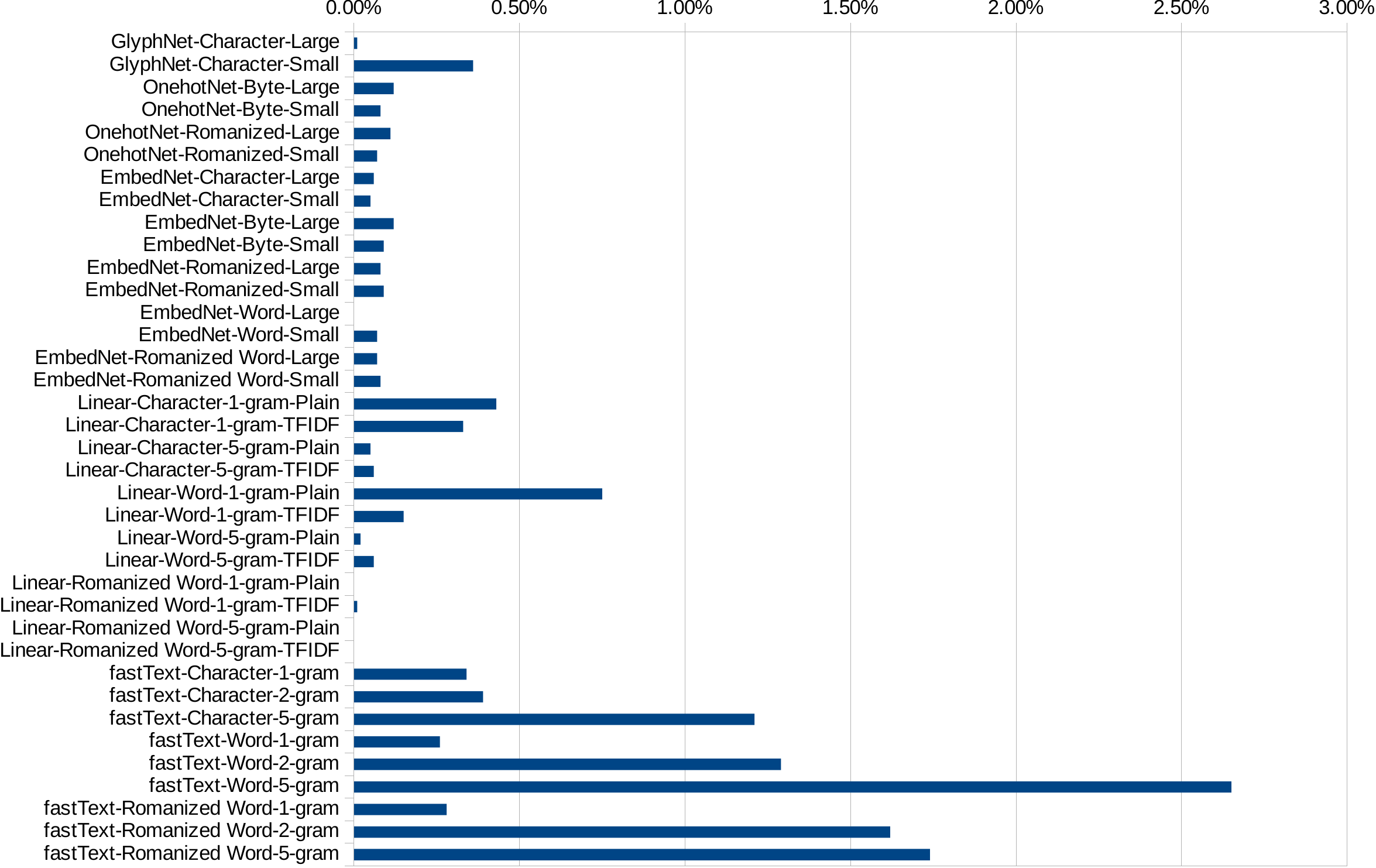}
  \end{center}
  \caption{Generalization gap of Joint binary dataset}
  \label{fig:gene}
\end{figure}

The optimizaiton algorithm used for all linear models is parallelized SGD. Each model is trained with a sparse representation via HOGWILD! \citep{NRRW11} parallelization using 10 CPU cores. An extra core is used for continuously testing on both training and testing datasets. The learning rate used for the algorithm is 0.001. A small weight decay of 0.00001 is applied to each model to stabilize the training process. The training stops after 1000 continuous testing steps are done. All of our models are run with a batch of INTEL XEON E5-2630 v2 CPUs.

The optimization parameters for fastText \citep{JGBM16} are controlled by the original authors' program \footnote{\url{https://github.com/facebookresearch/fastText}}. We set the embedding dimension to be 10 with a bucket size of 10,000,000, and going through each dataset for 2, 5 or 10 epoches depending on the validation result from 10\% of the training dataset. The optimization algorithm is SGD with decaying learning rate, where the initial learning rate is set to 0.1 and the decay change rate set to 100. The number of CPU cores used is 10, with a batch of INTEL XEON E5-2630 v2 CPUs. All other parameters used are the program's defaults.

\begin{figure}[t]
  \begin{center}
    \includegraphics[width=\textwidth]{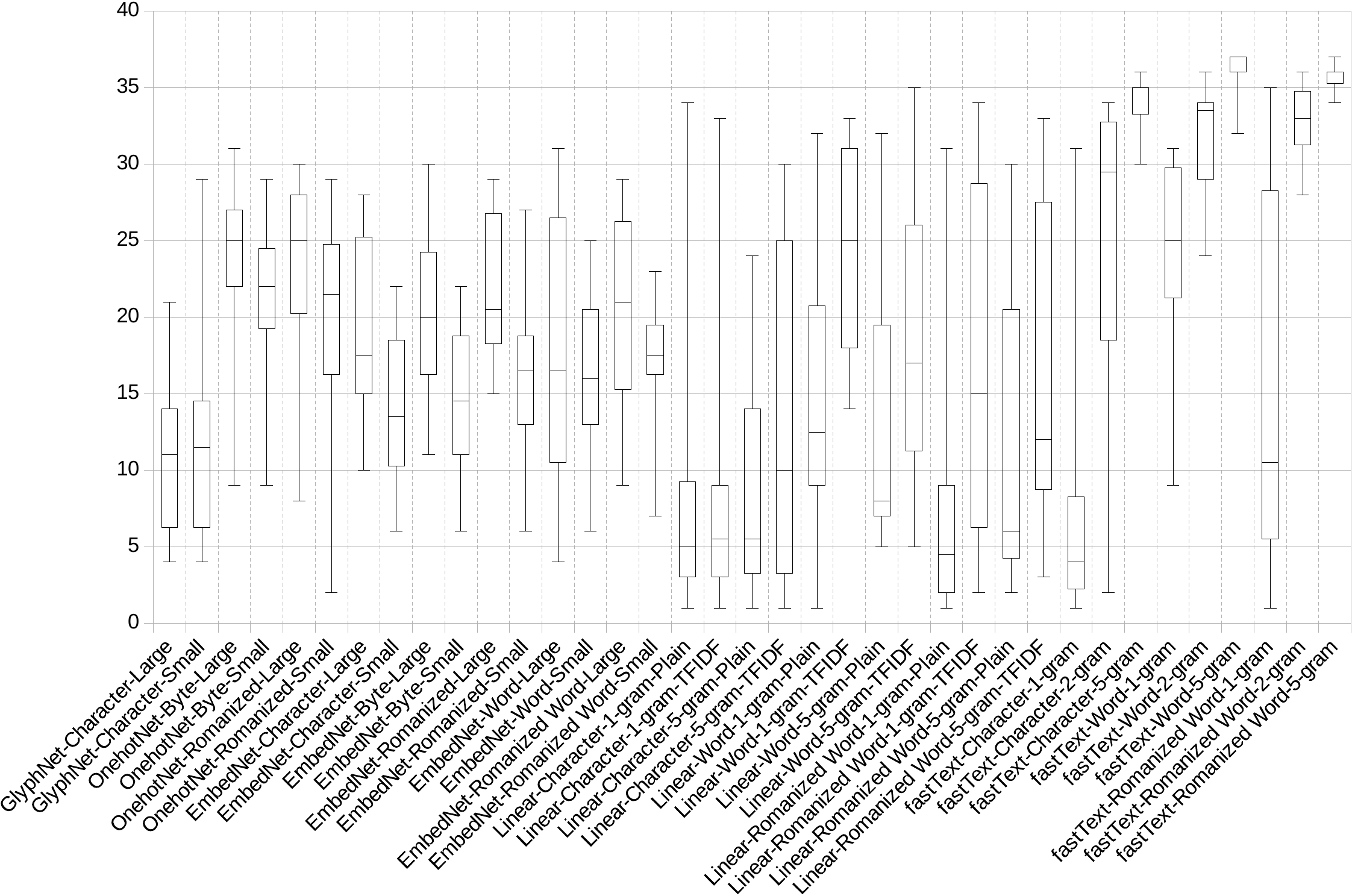}
  \end{center}
  \caption{Rank box plot of generalization gap for different models}
  \label{fig:rang}
\end{figure}

\subsection{Results}

The results for all the models are split into several tables. Table \ref{tab:glyr} lists the results for GlyphNet, where the numbers are testing errors in percentages. Similarly, Tables \ref{tab:oner}, \ref{tab:embr}, \ref{tab:linr} and \ref{tab:fstr} list the testing errors for OnehotNet, EmbedNet, linear models and fastText. As long as it is appicable in each table, the best result for each dataset is marked blue and the worst red. The epoch numbers for fastText models are presented in Appendix \ref{app:epft}

For each Chinese, Japanese and Korean dataset, we have 37 models each, and for English we have 22. In total, there are 473 models benchmarked in this article. Due to space limitations, the training errors are not present in the main text of this article, but readers can refer to Appendix \ref{app:trer} for them.

\begin{table}[t]
  \begin{center}
    \begin{tabular}{|l|l|l|r|r|}
      \hline
      \multicolumn{1}{|c|}{Model}  & \multicolumn{1}{c|}{Levels} & \multicolumn{1}{c|}{Variant} & \multicolumn{1}{c|}{Time} & \multicolumn{1}{c|}{Day-hh:mm:ss} \\ \hline
      \multirow{2}{*}{GlyphNet} & \multirow{2}{*}{Character} & Large & 136,250 & 1-13:50:50 \\ \cline{3-5}
      & & Small & 63,125 & 17:32:05 \\ \hline
      \multirow{2}{*}{OnehotNet} & \multirow{2}{*}{\begin{tabular}{@{}l@{}} Byte \\ Romanized \end{tabular}} & Large & 5,906 & 1:38:26 \\ \cline{3-5}
      & & Small & 5,331 & 1:28:51 \\ \hline
      \multirow{6}{*}{EmbedNet} & \multirow{2}{*}{Character} & Large & 2,143 & 35:43 \\ \cline{3-5}
      & & Small & 1,599 & 26:39 \\ \cline{2-5}
      & \multirow{2}{*}{\begin{tabular}{@{}l@{}} Byte \\ Romanized \end{tabular}} & Large & 1,829 & 30:29 \\ \cline{3-5}
      & & Small & 1,417 & 23:37 \\ \cline{2-5}
      & \multirow{2}{*}{\begin{tabular}{@{}l@{}} Word \\ Romanized Word \end{tabular}} & Large & 1,844 & 30:44 \\ \cline{3-5}
      & & Small & 1,536 & 25:36 \\ \hline
      \multirow{6}{*}{Linear} & \multirow{2}{*}{Character} & 1-gram & 1,518 & 25:18 \\ \cline{3-5}
      & & 5-gram & 7,647 & 2:07:27 \\ \cline{2-5}
      & \multirow{2}{*}{Word} & 1-gram & 1,417 & 23:37 \\ \cline{3-5}
      & & 5-gram & 6,250 & 1:44:10 \\ \cline{2-5}
      & \multirow{2}{*}{Romanized Word} & 1-gram & 1,333 & 22:13 \\ \cline{3-5}
      & & 5-gram & 6,253 & 1:44:13 \\ \hline
      \multirow{9}{*}{fastText} & \multirow{2}{*}{Character} & 1-gram & 7 & 7 \\ \cline{3-5}
      & & 2-gram & 7 & 7 \\ \cline{3-5}
      & & 5-gram & 10 & 10 \\ \cline{2-5}
      & \multirow{2}{*}{Word} & 1-gram & 2 & 2 \\ \cline{3-5}
      & & 2-gram & 3 & 3 \\ \cline{3-5}
      & & 5-gram & 5 & 5 \\ \cline{2-5}
      & \multirow{2}{*}{Romanized Word} & 1-gram & 3 & 3 \\ \cline{3-5}
      & & 2-gram & 3 & 3 \\ \cline{3-5}
      & & 5-gram & 5 & 5 \\ \hline
    \end{tabular}
  \end{center}
  \caption{Estimated training time for going over 1,000,000 samples using Joint binary dataset. The time estimation in the fourth column is in seconds. Encoding levels that will give the identical models are grouped together because the time estimation would be the same. These estimations are only for reference and may vary depending on actual computing environment.}
  \label{tab:time}
\end{table}

\section{Analysis}

In this section, we provide some analysis on the testing results presented in the previous section. These analyses include average ranks between models, generalization ability of each model under different encoding mechanisms, and estimations of training time.

\begin{figure}[t]
  \begin{center}
    \includegraphics[width=\textwidth]{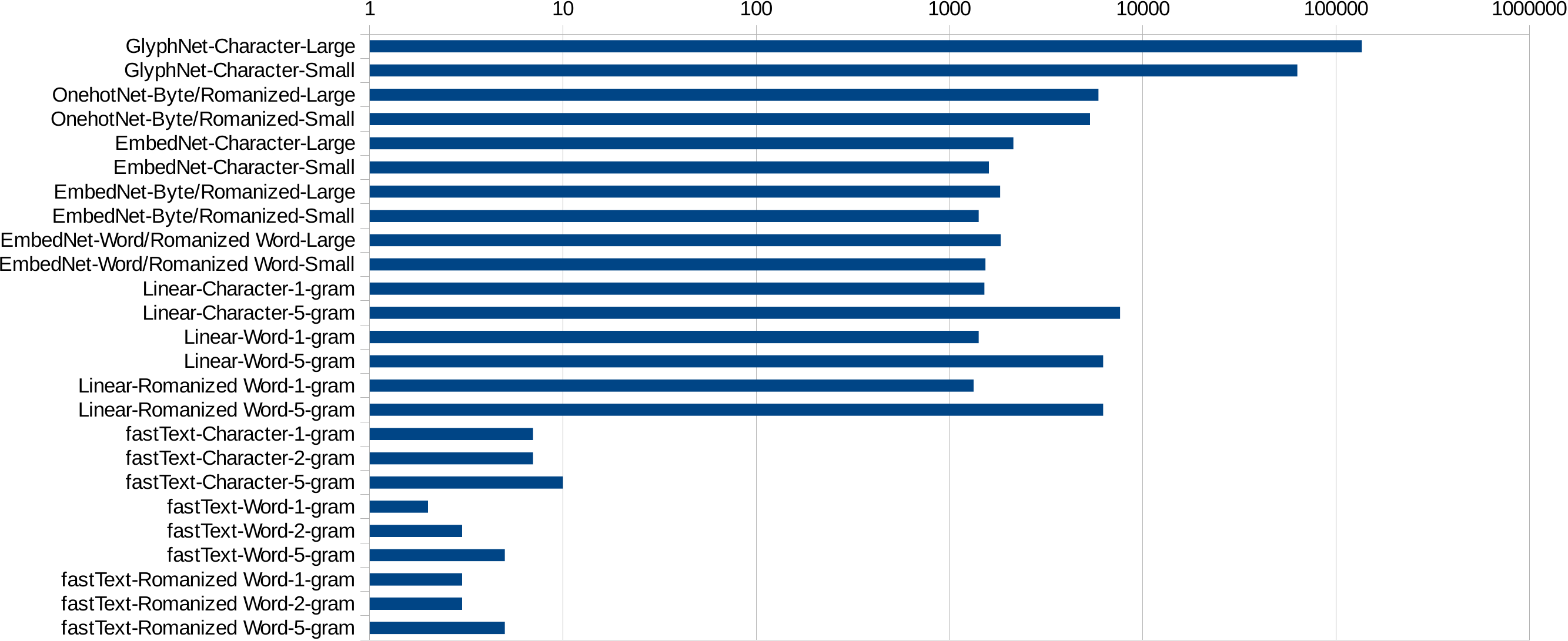}
  \end{center}
  \caption{Time for different models to go over 1,000,000 samples. The time axis is in logarithmic scale.}
  \label{fig:time}
\end{figure}

\subsection{Rank the Models}

To compare between different encoding mechanisms, this section presents the ranking of testing errors of all models. For English datasets, there are some missing values in various models in Tables \ref{tab:oner}, \ref{tab:embr}, \ref{tab:linr} and \ref{tab:fstr}. These values are missing because the corresponding models operate on romanized texts, and there is no romanization because the texts are already in the English alphabet. However, in order to make the model rank between different datasets comparable, we need to make sure that every dataset has the same number of models. To do this, we simply fill the missing values in romanized models with their corresponding ones for English. As a result, all datasets have 37 models to rank.

For each dataset, we rank all of the models in ascending order of their testing errors. The rank is the index of the model in this ordering. As a result, the smaller the rank, the better the model performs. Then, we compute the minimum, first quartile, median, third quartile, and maximum rank across different datasets for each model and put these numbers as a box plot in Figure \ref{fig:rank}. The numbers in Figure \ref{fig:rank} indicate both how each models perform on average, and how stable these models are across different datasets and languages.

From the results, the model achieved the best consistent performance is the character-level 5-gram fastText \citep{JGBM16} model. The result is more apparent in table \ref{tab:fstr}, where for almost all Chinese, Japanese and Korean datasets the best encoding is character-level 5-gram for fastText. For English, the best encoding is often word n-grams, although character-level 5-gram models are quite competitive as well. Character-level encoding with number of grams less than 5 are significantly worse, with the worst being bag-of-character linear model with TFIDF features. Word-level n-grams feature for both linear models and fastText are competitive, although our data processing pipeline did not guarantee perfect word segmentation for CJK languages because of the segmenters used.

\begin{wraptable}{r}{0.6\textwidth}
  \begin{center}
    \setlength\tabcolsep{2pt}
    \begin{tabular}{|l|r|r|r|r|r|r|}
      \hline
      \multicolumn{1}{|c|}{\multirow{3}{*}{Dataset}} & \multicolumn{2}{c|}{GlyphNet} & \multicolumn{4}{c|}{OnehotNet} \\ \cline{2-7} 
      & \multicolumn{2}{c|}{Character} & \multicolumn{2}{c|}{Byte} & \multicolumn{2}{c|}{Romanized} \\ \cline{2-7}
      & \multicolumn{1}{c|}{large} & \multicolumn{1}{c|}{small} & \multicolumn{1}{c|}{large} & \multicolumn{1}{c|}{small} & \multicolumn{1}{c|}{large} & \multicolumn{1}{c|}{small} \\
      \hline
      Dianping & 23.97 & 24.33 & 22.42 & 22.68 & 22.78 & 22.97 \\
      JD f. & 48.63 & 49.03 & 47.58 & 47.62 & 47.79 & 47.95 \\
      JD b. & 9.73 & 9.97 & 9.03 & 9.06 & 9.16 & 9.17 \\
      Rakuten f. & 46.59 & 46.94 & 44.62 & 44.95 & 44.74 & 44.96 \\
      Rakuten b. & 6.45 & 6.63 & 5.54 & 5.71 & 5.64 & 5.75 \\
      11st f. & 31.73 & 32.09 & 28.84 & 29.54 & 29.35 & 29.99 \\
      11st b. & 13.78 & 13.88 & 13.01 & 13.07 & 13.15 & 13.20 \\
      Amazon f. & 46.29 & 47.70 & 41.51 & 41.65 & \multicolumn{1}{|c|}{--} & \multicolumn{1}{|c|}{--} \\
      Amazon b. & 8.26 & 8.90 & 6.06 & 6.11 & \multicolumn{1}{|c|}{--} & \multicolumn{1}{|c|}{--} \\
      Ifeng & 16.35 & 17.35 & 12.43 & 13.62 & 14.99 & 15.90 \\
      Chinanews & 11.58 & 12.44 & 9.32 & 9.61 & 10.12 & 10.52 \\
      NYTimes & 17.24 & 17.99 & 12.36 & 12.57 & \multicolumn{1}{|c|}{--} & \multicolumn{1}{|c|}{--} \\
      Joint f. & 45.04 & 45.62 & 43.18 & 43.34 & 43.56 & 43.69 \\
      Joint b. & 9.97 & 10.04 & 8.67 & 8.70 & 8.89 & 8.95 \\
      \hline
    \end{tabular}
  \end{center}
  \caption{GlyphNet and OnehotNet training errors}
  \label{tab:gotr}
\end{wraptable}

Convolutional networks consistently have the best stability across different datasets and languages, with the best being byte-level large OnehotNet. This suggests that handling different language at byte-level regardless of whether characters could span multiple bytes is quite a feasible solution for handling different languages in a unified fashion. What is better is that byte-level language processing requires the least amount of pre-processing -- just present UTF-8 encoded strings to the model. Therefore, we believe byte-level model is a promising approach towards applying deep learning to natural language processing.

Finally, many different models have hit rank 1 as their minimum, suggesting that there is no single best models across different datasets and languages. However, this is limited to the model hyperparameters we chose. It is worth noting that hyperparameters are more thoroughly explored for fastText than other models in this article.

\subsection{Generalization}

In this section, we look at the generalization gap -- the expected difference between training and testing errors -- of different models. The generalization gap in this article is approximated by the subtraction of the training error from the testing error. The approximation to the underlying sample distribution should be pretty accurate because all our datasets are very large.

As an example, Figure \ref{fig:gene} visualizes the generalization gap for the Joint binary dataset. This figure exemplifies typical generalization properties of different models for all of our datasets. Additionally, Figure \ref{fig:rang} offers a box plot for the rankings on generalization error, computed in the same way as Figure \ref{fig:rank} for testing errors.

From these figures, one could easily observe that fastText \citep{JGBM16} tends to overfit much more aggressively than either convolutional networks or our own implementation of linear models, in spite of our effort in hyper-parameter tuning. Also, it overfits more using richer features as the number of grams goes from 1 to 5. Given the theoretical fact that fastText could not have more representation capacity than a linear model, this could be a result of the lack of regularization and the aggressive optimization strategy in fastText.

\begin{table}[t]
  \begin{center}
    \setlength\tabcolsep{2pt}
    \begin{tabular}{|l|r|r|r|r|r|r|r|r|r|r|}
      \hline
      \multicolumn{1}{|c|}{\multirow{2}{*}{Dataset}} & \multicolumn{2}{c|}{Character} & \multicolumn{2}{c|}{Byte} & \multicolumn{2}{c|}{Romanized} & \multicolumn{2}{c|}{Word} & \multicolumn{2}{c|}{Rom. word} \\ \cline{2-11}
      & \multicolumn{1}{c|}{large} & \multicolumn{1}{c|}{small} & \multicolumn{1}{c|}{large} & \multicolumn{1}{c|}{small} & \multicolumn{1}{c|}{large} & \multicolumn{1}{c|}{small} & \multicolumn{1}{c|}{large} & \multicolumn{1}{c|}{small} & \multicolumn{1}{c|}{large} & \multicolumn{1}{c|}{small} \\
      \hline
      Dianping & 22.97 & 23.17 & 23.33 & 23.60 & 24.66 & 25.45 & 24.03 & 24.25 & 23.07 & 23.31 \\
      JD f. & 47.80 & 47.90 & 47.99 & 48.22 & 48.15 & 48.74 & 49.50 & 49.68 & 48.58 & 48.65 \\
      JD b. & 9.23 & 9.24 & 8.96 & 9.04 & 9.17 & 9.48 & 10.19 & 10.23 & 9.34 & 9.47 \\
      Rakutenf f. & 44.85 & 45.40 & 45.60 & 46.55 & 45.69 & 45.59 & 46.03 & 46.24 & 45.71 & 46.10 \\
      Rakuten b. & 5.71 & 5.84 & 6.12 & 6.55 & 6.19 & 6.67 & 6.45 & 6.55 & 6.22 & 6.37 \\
      11st f. & 29.80 & 30.39 & 31.65 & 32.90 & 32.50 & 33.74 & 33.50 & 34.70 & 38.79 & 39.74 \\
      11st b. & 13.18 & 13.29 & 13.02 & 13.27 & 13.21 & 13.50 & 13.92 & 14.01 & 17.38 & 17.47 \\
      Amazon f. & 43.05 & 43.76 & \multicolumn{1}{c|}{--} & \multicolumn{1}{c|}{--} & \multicolumn{1}{c|}{--} & \multicolumn{1}{c|}{--} & 43.78 & 44.32 & \multicolumn{1}{c|}{--} & \multicolumn{1}{c|}{--} \\
      Amazon b. & 6.68 & 7.20 & \multicolumn{1}{c|}{--} & \multicolumn{1}{c|}{--} & \multicolumn{1}{c|}{--} & \multicolumn{1}{c|}{--} & 7.56 & 7.68 & \multicolumn{1}{c|}{--} & \multicolumn{1}{c|}{--} \\
      Ifeng & 13.71 & 14.44 & 14.44 & 15.36 & 16.82 & 18.17 & 16.85 & 17.72 & 15.66 & 16.65 \\
      Chinanews & 9.72 & 9.92 & 9.57 & 9.90 & 10.53 & 11.93 & 14.75 & 13.68 & 10.56 & 10.86 \\
      NYTimes & 12.75 & 13.51 & \multicolumn{1}{c|}{--} & \multicolumn{1}{c|}{--} & \multicolumn{1}{c|}{--} & \multicolumn{1}{c|}{--} & 15.83 & 16.15 & \multicolumn{1}{c|}{--} & \multicolumn{1}{c|}{--} \\
      Joint f. & 43.74 & 44.18 & 44.35 & 44.94 & 44.88 & 45.70 & 45.33 & 45.62 & 45.15 & 45.45 \\
      Joint b. & 8.96 & 9.13 & 8.97 & 9.18 & 9.26 & 9.56 & 10.75 & 9.94 & 9.87 & 9.94 \\
      \hline
    \end{tabular}
  \end{center}
  \caption{EmbedNet training errors}
  \label{tab:embt}
\end{table}

However, the fact that models with simpler representation capacity can overfit so aggressively indicates that generalization does not only depend on the complexity of the model or the number of parameters in the model, but also its capacity to represent the data for the task at hand. This aspect may be the reason why on average models like convolutional networks can achieve much better results than what can be characterized by the upper-bounds of traditional learning theory. This requires further study beyond the current generalization bounds based on statistical concentration inequalities and complexity measurements, and it may require a better characterization between the relationship of representation and generalization.

\subsection{Training Time}

The training times of different models vary greatly in our experiments. Table \ref{tab:time} offers an estimation of time it took for each model to go over 1,000,000 samples with the hardware mentioned in the previous section. In general, fastText \citep{JGBM16} offers the best training time and only requires CPUs, whereas convolutional networks take the longest time and require GPUs. Depending on the methods of encoding, the performance between convolutional networks also differ drastically, with EmbedNet tens of times faster than GlyphNet. Figure \ref{fig:time} visualizes the estimations as a bar chart.

These results show that fastText \citep{JGBM16} offers the fastest training and evaluation while achieving competitive results. On the other hand, models using convolutional networks consume the most amount of computation time. As a result, in this article we could afford to do hyper-parameter tuning for fastText but not on convolutional networks.

The convolutional network models in this article are designed not for achieving the best performance, but for the fairness of comparing between different encoding mechanisms within the computational budget we possess. Given the fact that different designs of convolutional networks could offer drastically different performance, we believe there is a great deal of potential for improvement from different design choices on convolutional networks.

\begin{table}[t]
  \begin{center}
    \setlength\tabcolsep{2pt}
    \begin{tabular}{|l|r|r|r|r|r|r|r|r|r|r|r|r|}
      \hline
      \multicolumn{1}{|c|}{\multirow{3}{*}{Dataset}}  & \multicolumn{4}{c|}{Character} & \multicolumn{4}{c|}{Word} & \multicolumn{4}{c|}{Romanized Word} \\ \cline{2-13}
      & \multicolumn{2}{c|}{1-gram} & \multicolumn{2}{c|}{5-gram} & \multicolumn{2}{c|}{1-gram} & \multicolumn{2}{c|}{5-gram} & \multicolumn{2}{c|}{1-gram} & \multicolumn{2}{c|}{5-gram} \\ \cline{2-13}
      & \multicolumn{1}{c|}{plain} & \multicolumn{1}{c|}{tfidf} & \multicolumn{1}{c|}{plain} & \multicolumn{1}{c|}{tfidf} & \multicolumn{1}{c|}{plain} & \multicolumn{1}{c|}{tfidf} & \multicolumn{1}{c|}{plain} & \multicolumn{1}{c|}{tfidf} & \multicolumn{1}{c|}{plain} & \multicolumn{1}{c|}{tfidf} & \multicolumn{1}{c|}{plain} & \multicolumn{1}{c|}{tfidf} \\
      \hline
      Dianping & 26.03 & 26.72 & 24.26 & 23.30 & 23.94 & 23.39 & 23.47 & 22.59 & 27.35 & 28.03 & 24.37 & 23.14 \\
      JD f. & 50.84 & 51.30 & 47.27 & 46.08 & 47.74 & 46.55 & 45.86 & 43.62 & 52.19 & 52.61 & 47.74 & 46.15 \\
      JD b. & 11.84 & 12.12 & 8.99 & 8.68 & 9.68 & 9.49 & 8.89 & 8.28 & 13.06 & 13.35 & 9.09 & 8.75 \\
      Rakutenf f. & 52.21 & 52.82 & 47.18 & 45.57 & 46.96 & 45.90 & 45.55 & 43.61 & 47.66 & 46.71 & 46.85 & 44.17 \\
      Rakuten b. & 12.43 & 12.94 & 8.10 & 7.25 & 8.35 & 8.13 & 7.19 & 6.47 & 8.78 & 8.71 & 7.31 & 6.57 \\
      11st f. & 43.51 & 47.63 & 43.14 & 43.16 & 44.14 & 42.20 & 42.16 & 40.62 & 40.17 & 35.30 & 40.52 & 35.29 \\
      11st b. & 17.73 & 18.01 & 14.33 & 14.24 & 15.09 & 15.03 & 13.34 & 12.91 & 14.32 & 13.87 & 13.83 & 12.79 \\
      Amazon f. & 69.28 & 68.13 & 56.28 & 50.01 & 44.84 & 42.85 & 43.97 & 41.78 & \multicolumn{1}{c|}{--} & \multicolumn{1}{c|}{--} & \multicolumn{1}{c|}{--} & \multicolumn{1}{c|}{--} \\
      Amazon b. & 34.36 & 33.91 & 14.92 & 12.08 & 9.15 & 8.37 & 8.43 & 8.03 & \multicolumn{1}{c|}{--} & \multicolumn{1}{c|}{--} & \multicolumn{1}{c|}{--} & \multicolumn{1}{c|}{--} \\
      Ifeng & 22.00 & 21.78 & 21.42 & 21.35 & 18.49 & 16.51 & 19.60 & 18.34 & 26.59 & 26.81 & 23.08 & 22.06 \\
      Chinanews & 15.16 & 14.93 & 15.14 & 13.31 & 11.28 & 10.00 & 13.32 & 12.68 & 20.10 & 20.39 & 15.62 & 13.76 \\
      NYTimes & 57.21 & 53.77 & 39.78 & 26.29 & 17.81 & 14.30 & 19.63 & 17.88 & \multicolumn{1}{c|}{--} & \multicolumn{1}{c|}{--} & \multicolumn{1}{c|}{--} & \multicolumn{1}{c|}{--} \\
      Joint f. & 57.16 & 56.72 & 49.68 & 47.43 & 46.53 & 45.57 & 44.97 & 44.43 & 47.93 & 47.19 & 47.22 & 46.61 \\
      Joint b. & 19.77 & 19.40 & 12.07 & 10.85 & 10.08 & 10.51 & 9.44 & 8.94 & 11.86 & 11.43 & 11.41 & 11.02 \\
      \hline
    \end{tabular}
  \end{center}
  \caption{Linear model training errors}
  \label{tab:lint}
\end{table}

It is also worth noting that the task in question -- text classification -- is quite simple. Convolutional networks may not show an advantage in this specific task, but may become more useful for more complicated reasoning tasks concerning text inputs and outputs. The comparison between different encoding mechanisms presented this article offer valuable knowledge towards the choice for convolutional networks in general language processing.

\section{Other Models}

In spite of the 473 models we have benchmarked, this article is in no way a complete essay on every possible model for text classification. Some of the interesting models we did not benchmark include recurrent networks, the use of sparse convolutions for text, and different variations of convolutional architectures.

By focusing on different encoding mechanisms for deep learning models, this article performs experiments only on one kind -- convolutional networks. Another often-used kind for processing texts is recurrent networks, constructed using different types of cells like long short-term memory (LSTM) \citep{HS97} and gated recurrent units (GRU) \citep{CMBB14}. Some authors have found that recurrent networks applied to different levels of encoding can offer good results for text classification as well (for example, \citet{DL15} and \citet{LQH16}). Combinations of convolutional networks and recurrent networks are also explored for text classification (for example, \citet{XC16}).

\begin{table}[t]
  \begin{center}
    \setlength\tabcolsep{2pt}
    \begin{tabular}{|l|r|r|r|r|r|r|r|r|r|}
      \hline
      \multicolumn{1}{|c|}{\multirow{2}{*}{Dataset}} & \multicolumn{3}{c|}{Character} & \multicolumn{3}{c|}{Word} & \multicolumn{3}{c|}{Romanized Word} \\ \cline{2-10}
      & \multicolumn{1}{c|}{1-gram} & \multicolumn{1}{c|}{2-gram} & \multicolumn{1}{c|}{5-gram} & \multicolumn{1}{c|}{1-gram} & \multicolumn{1}{c|}{2-gram} & \multicolumn{1}{c|}{5-gram} & \multicolumn{1}{c|}{1-gram} & \multicolumn{1}{c|}{2-gram} & \multicolumn{1}{c|}{5-gram} \\
      \hline
      Dianping & 25.82 & 21.03 & 19.34 & 23.12 & 21.45 & 16.21 & 27.07 & 22.07 & 19.43 \\
      JD f. & 50.98 & 47.36 & 44.34 & 48.04 & 46.14 & 39.51 & 51.97 & 46.58 & 43.32 \\
      JD b. & 11.83 & 8.32 & 7.49 & 9.60 & 6.19 & 6.14 & 13.01 & 8.64 & 7.48 \\
      Rakutenf f. & 51.92 & 43.38 & 40.54 & 45.83 & 42.30 & 36.13 & 46.57 & 41.15 & 37.66 \\
      Rakuten b. & 12.15 & 6.51 & 3.91 & 7.80 & 4.88 & 3.26 & 8.30 & 5.18 & 3.68 \\
      11st f. & 42.94 & 35.91 & 25.96 & 39.08 & 35.49 & 30.93 & 39.06 & 31.94 & 25.74 \\
      11st b. & 17.68 & 13.17 & 12.12 & 15.12 & 12.75 & 10.63 & 11.80 & 11.29 & 8.11 \\
      Amazon f. & 67.00 & 53.88 & 38.40 & 42.30 & 37.60 & 31.15 & \multicolumn{1}{c|}{--} & \multicolumn{1}{c|}{--} & \multicolumn{1}{c|}{--} \\
      Amazon b. & 32.72 & 18.60 & 5.46 & 7.78 & 3.47 & 0.22 & \multicolumn{1}{c|}{--} & \multicolumn{1}{c|}{--} & \multicolumn{1}{c|}{--} \\
      Ifeng & 21.00 & 11.21 & 3.90 & 13.84 & 10.81 & 0.62 & 25.84 & 13.11 & 4.40 \\
      Chinanews & 13.55 & 5.93 & 0.17 & 7.03 & 1.61 & 0.02 & 18.58 & 6.53 & 2.36 \\
      NYTimes & 51.07 & 24.13 & 8.75 & 10.73 & 3.53 & 6.47 & \multicolumn{1}{c|}{--} & \multicolumn{1}{c|}{--} & \multicolumn{1}{c|}{--} \\
      Joint f. & 56.79 & 46.61 & 40.52 & 45.47 & 41.98 & 33.90 & 47.01 & 41.08 & 38.17 \\
      Joint b. & 19.51 & 11.96 & 7.79 & 10.31 & 7.36 & 6.09 & 11.11 & 7.19 & 7.10 \\
      \hline
    \end{tabular}
  \end{center}
  \caption{fastText training errors}
  \label{tab:fstt}
\end{table}

This article explores one-hot encoding for convolutional networks using byte-level encoding and romanization. Another alternative is to implement a convolutional module that can take sequences of indices instead of explicit vectors to represent one-hot encoding. This would avoid the memory overflow problem when applying one-hot encoding to large vocabularies. However, so far there has been no deep learing toolkit that has implementation of such a sparse convolutional module. Furthermore, it may require special numerical optimization that would merit its own essay. Therefore, it is not included for presentation in this article.

Finally, the results on convolutional networks in this article are limited to the purpose of offering fair comparisons between different encoding mechanisms. Another dimension of exploration is the design variants of convolutional networks for text processing, such as very deep networks \citep{CSBL17}, residual \citep{HZRS16} and dense \citep{HLWKM16} connections, and advanced pooling schemes for handling the variable length problem \citep{KGB14} \citep{JZ17}. We are optimistic that exploration of all these different architecture designs could improve the results further for convolutional networks.

\section{Conclusion}

This article explores the use of different encoding mechanisms for both deep learning and linear models for text classification in Chinese, English, Japanese and Korean. These encoding mechanisms include one-hot encoding, embedding and images of character glyphs. Different levels of encoding are applied to each mechanism whenever application, including UTF-8 encoded bytes, characters, words, romanized characters and romanized words. There are in total 473 models benchmarked in this article, including convolutional networks, linear models and fastText \citep{JGBM16}.

A total of 14 large-scale datasets were built in this article for benchmarking these models in 4 languages including Chinese, English, Japanese and Korean. Most of these datasets have millions of samples for training, and 2 of these datasets contains samples mixed in all these 4 languages to testing different model's ability to handle different languages in a consistent and unified fashion.

\begin{table}[t]
  \begin{center}
    \setlength\tabcolsep{2pt}
    \begin{tabular}{|l|c|c|c|c|c|c|c|c|c|}
      \hline
      \multicolumn{1}{|c|}{\multirow{2}{*}{Dataset}} & \multicolumn{3}{c|}{Character} & \multicolumn{3}{c|}{Word} & \multicolumn{3}{c|}{Romanized Word} \\ \cline{2-10}
      & \multicolumn{1}{c|}{1-gram} & \multicolumn{1}{c|}{2-gram} & \multicolumn{1}{c|}{5-gram} & \multicolumn{1}{c|}{1-gram} & \multicolumn{1}{c|}{2-gram} & \multicolumn{1}{c|}{5-gram} & \multicolumn{1}{c|}{1-gram} & \multicolumn{1}{c|}{2-gram} & \multicolumn{1}{c|}{5-gram} \\
      \hline
      Dianping & 10 & 10 & 2 & 10 & 2 & 2 & 5 & 5 & 2 \\
      JD f. & 2 & 2 & 2 & 5 & 2 & 2 & 10 & 5 & 2 \\
      JD b. & 10 & 10 & 2 & 10 & 10 & 2 & 10 & 5 & 2 \\
      Rakutenf f. & 10 & 10 & 2 & 10 & 2 & 2 & 10 & 5 & 2 \\
      Rakuten b. & 10 & 10 & 5 & 10 & 5 & 2 & 10 & 5 & 2 \\
      11st f. & 5 & 5 & 5 & 5 & 2 & 2 & 2 & 2 & 2 \\
      11st b. & 10 & 10 & 2 & 2 & 2 & 2 & 5 & 2 & 2 \\
      Amazon f. & 10 & 10 & 10 & 10 & 2 & 2 & \multicolumn{1}{c|}{--} & \multicolumn{1}{c|}{--} & \multicolumn{1}{c|}{--} \\
      Amazon b. & 10 & 10 & 10 & 10 & 5 & 5 & \multicolumn{1}{c|}{--} & \multicolumn{1}{c|}{--} & \multicolumn{1}{c|}{--} \\
      Ifeng & 10 & 5 & 5 & 5 & 2 & 5 & 10 & 5 & 5 \\
      Chinanews & 10 & 5 & 10 & 5 & 5 & 10 & 10 & 5 & 5 \\
      NYTimes & 10 & 10 & 10 & 5 & 5 & 2 & \multicolumn{1}{c|}{--} & \multicolumn{1}{c|}{--} & \multicolumn{1}{c|}{--} \\
      Joint f. & 10 & 10 & 2 & 10 & 2 & 2 & 10 & 5 & 2 \\
      Joint b. & 10 & 10 & 2 & 10 & 5 & 2 & 10 & 10 & 2 \\
      \hline
    \end{tabular}
  \end{center}
  \caption{fastText epoches}
  \label{tab:fstp}
\end{table}

Some conclusions from these results are:

\begin{enumerate}
\item fastText \citep{JGBM16} has the best result with character-level n-gram encoding for Chinese, Japanese and Korean texts. For English, the best encoding for fastText is word-level n-grams. 
\item Word-level encoding for CJK languages are competitive even without perfect segmentation, for both fastText and linear models.
\item The best encoding mechanism for convolutional networks is byte-level one-hot encoding. This indicates that convolutional networks have the ability to understand text from a low-level representation, and offers great simplicity for handling multiple languages in a consistent and unified fashion.
\item fastText tends to overfit more than convolutional networks, in spite of the fact that it does not have more representation capacity than a linear model.
\end{enumerate}

In the future, we hope to extend the results to recurrent networks, and explore how different designs of convolutional networks would affect the results. We plan to release all the source code used for all the benchmarks, and hope that these results are useful for the community to choose which encoding mechanism to use when facing with multi-lingual text processing.

\section*{Acknowledgement}

We want to express our thanks to Junyoung Chung, Yoon Kim, Sainbayar Sukhbaatar and Kentaro Hanaki for offering their knowledge on Korean and Japanese languages. Armand Joulin offered valuable suggestions on hyper-parameter tuning for fastText.

\appendix

\section{Training Errors of All Models}
\label{app:trer}

The training errors of all models are detailed in Tables \ref{tab:gotr}, \ref{tab:embt}, \ref{tab:lint} and \ref{tab:fstt}.

\section{Epoches for fastText Models}
\label{app:epft}

The validated epoches for running all fastText models are detailed in Table \ref{tab:fstp}.

\bibliography{article}{}

\end{document}